\documentclass[a4paper, 10pt, conference]{IEEEtran}      
\usepackage{FG2017}

\usepackage{lipsum}

\FGfinalcopy 

\IEEEoverridecommandlockouts                              

\usepackage[pdftex]{graphicx}
\usepackage{arydshln}
\usepackage{colortbl}

\title{\LARGE \bf
Local Shape Spectrum Analysis for 3D Facial Expression Recognition
}

\author{\parbox{16cm}{\centering
   {\large Dmytro Derkach and Federico M. Sukno}\\ 
    {\normalsize
    Department of Information and Communication Technologies, Pompeu Fabra University, Barcelona, Spain\\
    }}
}

\begin{document}

\IEEEoverridecommandlockouts\IEEEpubid{\makebox[\columnwidth]{978-1- 5090-4023-0/17/\$31.00~\copyright{}2017 IEEE \hfill} \hspace{\columnsep}\makebox[\columnwidth]{ }}

\ifFGfinal
\thispagestyle{empty}
\pagestyle{empty}
\else
\author{Anonymous FG 2017 submission\\-- DO NOT DISTRIBUTE --\\}
\pagestyle{plain}
\fi
\maketitle

\bstctlcite{IEEEexample:BSTcontrol}

\begin{abstract}
We investigate the problem of facial expression recognition using 3D data. Building from one of the most successful frameworks for facial analysis using exclusively 3D geometry, we extend the analysis from a curve-based representation into a spectral representation, which allows a complete description of the underlying surface that can be further tuned to the desired level of detail. Spectral representations are based on the decomposition of the geometry in its spatial frequency components, much like a Fourier transform, which are related to intrinsic characteristics of the surface. In this work, we propose the use of Graph Laplacian Features (GLF), which results from the projection of local surface patches into a common basis obtained from the Graph Laplacian eigenspace. We test the proposed approach in the BU-3DFE database in terms of expressions and Action Units recognition. Our results confirm that the proposed GLF produces consistently higher recognition rates than the curves-based approach, thanks to a more complete description of the surface, while requiring a lower computational complexity. We also show that the GLF outperform the most popular alternative approach for spectral representation, Shape-DNA, which is based on the Laplace Beltrami Operator and cannot provide a stable basis that guarantee that the extracted signatures for the different patches are directly comparable.
\end{abstract}

\section{INTRODUCTION}
Human face plays an important role while expressing emotions such as happiness, satisfaction, surprise, fear, sadness or disgust. 
While there is consensus about the need to integrate multi-modal information for a complete understanding of human emotions, facial expressions are considered one of the most relevant channels for humans to regulate interactions both with the environment and with other persons \cite{pantic2009machine}.

During the past two decades, the problem of facial expression recognition has become very relevant. The growing interest in improving the interaction and cooperation between people and
computers makes it necessary that automatic systems are able to
react to a user and his emotions, as it takes place in natural human
intercourse. Many applications such as virtual reality,
video-conferencing, user profiling and customer satisfaction studies
for broadcast and web services, require efficient facial expression
recognition in order to achieve the desired results \cite{girard2013social,
mcduff2013predicting}. Therefore, the impact of facial expression recognition on
the above-mentioned application areas is constantly growing.

Methods for facial expression recognition are generally based on two
possible imaging domains: 2D and 3D. Previous studies have focused
primarily on the 2D domain (texture information) \cite{sandbach2012static} due
to the prevalence of data. With the rapid development of 3D imaging
and scanning technologies, it becomes more and more popular using 3D
face scans. Compared with 2D face images, 3D face scans contain
detailed geometric shape information of facial surfaces, which
remove the problems of illumination and pose variations that are
inherent to the 2D modality. Thus, 3D-shape analysis has attracted
increasing attention \cite{li2015efficient}.

The availability of 3D information is not always fully
exploited and, in many cases, 3D information is analyzed by directly
applying 2D techniques to limited depth representation. This is
typically done by using depth maps (2.5D representations), where
the depth information is treated analogously to a gray-scale image
and the 3D information is simply extracted by computing popular 2D
texture descriptors such as LBPs \cite{guo2009method, wang20133d,
wang20143d} or Gabor filters \cite{yun2010human, xie2010fusing, d2007precise}. Following a similar strategy, Zeng et al. \cite{zeng2013automatic} conformally mapped the 3D facial surface to a 2D unit disk and then considered it as a 2D image. More recently, deep convolutional neural networks has been explored in order to generate deep features \cite{li2015deep} from this 2.5D representation.

However, in order to take full advantage of depth information we need
approaches that are truly 3D. A notable approach in this direction,
from Klassen et al., is based on the representation of surfaces with
a finite number of \emph{level curves} \cite{klassen2006geodesics}. They showed
that curves can be used to represent surface regions, being able to
capture quite subtle deformations. Thus, 3D shape analysis can be
performed by comparisons of corresponding level curves. It should be noted, however, that such comparison is not trivial, given that
distances between 3D level curves should be computed based on the
geodesic paths of their underlying manifold. An important step
forward in this direction was presented by Srivastava et al.
\cite{srivastava2011shape}, who introduced a square-root velocity
representation for analyzing curves in Euclidean spaces under a
Riemannian metric. In particular, they computed geodesic paths between
curves under this metric to obtain deformations between closed
curves. Samir et al. \cite{samir2009intrinsic} applied this curves-based approach for the analysis of facial surfaces.
They represented a surface as an indexed collection of closed curves. These curves were extracted according to to their Euclidean distance from the tip of the nose, which is sensitive to deformations and, thus, can better capture differences related to variant expressions. Then, authors studied curves' differential geometry and endowed it with a Riemannian metric. In order to quantify
differences between any two facial surfaces, the length of a
geodesic was used. A similar framework was used in \cite{drira20133d,
amor20144} for analyzing 3D faces, with the goal of comparing, matching
and averaging faces, with the difference that surfaces were
represented by radial curves outflowing from the nose tip. Maalej et al. \cite{maalej2011shape}, based on an indexed collection of closed curves, emphasized the importance of using local regions instead of the entire face and proposed a local geometric analysis of the surface. They introduced a facial surfaces representation based on sets of level curves around landmarks. In their work, they used 70 landmarks and then extracted collections of closed curves using Euclidean distance. Thereby, 70 patches centered on the considered points represented the facial surface, where each patch consisted of an indexed collection of 3D closed curves. Further, they applied a Riemannian framework to derive 3D shape analysis and quantify similarity between corresponding patches on different 3D facial scans.

Despite the success of the level-curves framework, it could be argued that it is an incomplete representation of the 3D data, since it only captures part of the underlying surface, which is actually sampled by means of a finite number of curves.  Spectral representations are based on the decomposition of the geometry in its (fundamental) frequency components, which are related to intrinsic characteristics of the surface, and correspond to the eigenvectors of the Laplace Beltrami Operator (LBO). The spectrum of the LBO is an isometric invariant, and it has been shown to be a powerful descriptor as a signature for (non-rigid) 3D shape matching and classification \cite{karni2000spectral,reuter2006laplace}. The most popular of such descriptors was proposed by Reuter et al. \cite{reuter2006laplace}, by taking the eigenvalues (i.e. spectrum) of its Laplace-Beltrami
operator. Because such spectrum captures intrinsic shape information they called the method “Shape-DNA”. It was shown that this approach can be used (like DNA-test) to identify 3D objects or to detect similarities in practical applications. Several works used the Shape-DNA to identify objects for the purpose of copyright protection, but, to the best of our knowledge, it has not been applied for facial expression analysis.

\subsection*{Contributions} In this paper, we explore the use of spectral methods as local shape descriptors for 3D facial expression recognition. We show that the
application of Shape-DNA is not the best way to deal with local face patches and that a fixed-graph basis, which we refer to as Graph Laplacian Features (GLF), provides
superior results. This is theoretically sound given the impossibility to ensure a fixed ordering of the spectral components under the Shape-DNA approach \cite{jain2007non}. Compared to the curves-based framework, the proposed method constitutes a generalization to a full representation of the surface patches resulting in higher accuracy and reduced computational complexity. We perform experiments over the BU-3DFE database and show that the proposed GLF approach consistently outperforms the curves-based and Shape-DNA alternatives, both in terms of expression and Action Unit recognition. 

\section{SPECTRAL SHAPE ANALYSIS}

Spectral shape analysis relies on the decomposition of the surface geometry  into its spatial frequency components (spectrum). Such representation allows to analyse the surface by examining 
the eigenvalues, eigenvectors or eigenspace projections of these fundamental frequencies.

One of the advantages of these methods is that they are invariant with
respect to isometry, which means that these descriptors do not
change with different isometric embeddings of the shape. In
addition, their advantage is that they can be applied well to
deformable objects. Spectral methods have been applied to solve a
variety of problems including mesh compression, correspondence,
smoothing, watermarking, segmentation, surface reconstruction etc.
\cite{reuter2010hierarchical, nealen2006laplacian, zhang2010spectral}.

In our work, we use the spectrum based on Laplace operator for
facial expression recognition. The Laplacians are the most commonly
used operators for spectral mesh processing. As Chung stated in her
book \cite{chung1997spectral}, results from spectral theory suggest that the
Laplacian eigenvalues are tightly related to almost all major graph
invariants. Thus, if data models the structures of a shape, either
topology or geometry, then it is expected that its set of
eigenvalues provides an appropriate characterization of the shape.
The eigenvalues serve as compact global shape descriptor
\cite{zhang2010spectral}.

Several Laplacian operators have been proposed in the literature to
compute the mesh spectrum. In this work we are especially interested
in the two most popular ones:
\begin{enumerate}
\item Graph Laplacian, related to operators that have been widely studied in
graph theory \cite{chung1997spectral}. Despite this operator is based solely
upon topological information, its eigenfunctions (i.e. eigenvectors)
generally have a remarkable conformity to the mesh geometry
\cite{isenburg2001connectivity}. On the other hand, the eigenfunctions of this
operator are sensitive to aspects such as mesh resolution or
triangulation.
\item Discretizations of the Laplace-Beltrami operator from
Riemannian geometry \cite{chavel1984eigenvalues, rosenberg1997laplacian}, which try to make basis
dependant only on the underlying geometry and not its specific
representation. This is the type of operator used in the Shape-DNA
approach.
\end{enumerate}


\begin{figure*}[t]
    \noindent
    \centering{
    \includegraphics[scale=0.68]{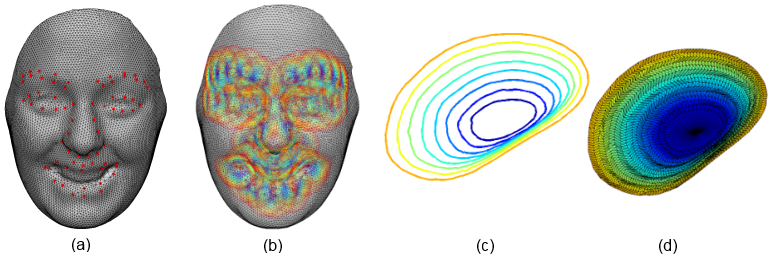} 
    }
    \caption{(a) 3D annotated facial shape model (68 landmarks); (b) closed curves extracted around the landmarks; (c) example of 8 level curves; (d) the mesh patche.}
    \label{3d}
\end{figure*}

\subsection{Graph Laplacian}

Mesh (graph) Laplacian operators are linear operators that act on
functions defined on the mesh and they depend purely on the mesh
points (vertices) and their connectivity (e.g. triangulation). Thus,
if mesh $M$ has $n$ vertices, a mesh Laplacian will be described by
a $n \times n$ matrix $L$.

Given a mesh $M$ with vertices $V$ and edges $E$, $M=(V, E)$, the graph Laplacian $L=L(M)$
is defined as
     $$L_{ij}=\left\{\begin{array}{ll}
        -1 & if ~(i,j)\in E \nonumber\\
        d_i & if ~i = j \\
        0 & otherwise \nonumber
    \end{array}\right.  \eqno(1)$$
where $d_i$ is the degree or valence of vertex $i$.

Since this operator is determined purely by the connectivity of the mesh,
it does not explicitly encode geometric information. However, 
as shown in the seminal work from Taubin \cite{taubin1995signal}, eigen-decomposition of the graph Laplacian produces an orthogonal basis whose components relate to spatial frequencies, much like a Fourier Transform. Projections of a mesh into the eigenspace of Laplacian operators have been proposed \cite{desbrun1999implicit, kim2005geofilter} and used to derive shape descriptors \cite{zahn1972fourier}.  In face, eigenvectors are most frequently used to derive a spectral embedding of the input data (e.g. the
mesh shape), since the spectral domain is more convenient to operate as it is low-dimensional and invariant to isometries while it still retains as much information about the input data as possible.
\subsection{Shape-DNA}
In Riemannian geometry, the Laplace operator can be generalized to operate on functions defined on a surface. In this case, the Laplace-Beltrami operator is of particular interest in geometry processing.

Ovsjanikov in \cite{ovsjanikov2008global} showed that the Laplace-Beltrami operator can be defined entirely in terms of the metric tensor on the manifold independently of the parametrization. Compared to the graph Laplacian, the Laplace-Beltrami operator does not operate on any mesh vertices, but rather on the underlying manifold itself. It depends continuously on the shape of the surface \cite{courant1965methods}.

\begin{figure}[b]
    \noindent
    \centering{
    \includegraphics[scale=0.5]{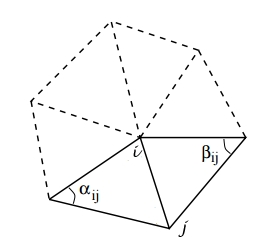}
    }
    \caption{1-ring neighbors and angles opposite to an edge.}
    \label{neighbors}
\end{figure}

The Laplace operator based on the cotan formula represents the most popular discrete approximation to the Laplace-Beltrami operator currently used for geometry processing. This operator can be presented as a product of a diagonal and symmetric matrix $L=B^{-1} S$. Where $B^{-1}$ is a diagonal matrix whose diagonal entries are Voronoi areas \cite{meyer2003discrete} for all vertices and $S$ is a symmetric matrix defined \cite{wang2012empirical}:

 $$S_{ij}=\left\{\begin{array}{ll}
       -w_{ij} & if ~(i,j)\in E  \nonumber\\
        \sum_{k \in N(i)}w_{ik}
        & if ~i=j \\
        0 & otherwise \nonumber
    \end{array}\right.  \eqno(2)$$
where $w_{ij}=(\cot\alpha_{ij}+\cot\beta_{ij})$, $\alpha_{ij}$ and $\beta_{ij}$ are the angles opposite if the edge $(i, j)$ (Fig. \ref{neighbors}). $N(i)$ is a set of vertices that are adjacent to vertixe $i$.

A significant amount of geometric and topological information is known to be contained in the spectrum. Since the spectrum (i.e. the eigenvalues) of the Laplace–Beltrami operator contains intrinsic shape information Reuter et al proposed to use them as shape signature or “Shape-DNA” \cite{reuter2006laplace}. Shape-DNA can be used to identify shapes and detect similarities.

In order to extract appropriate eigenvalues, matrix $L$ should be symmetric. The main advantage offered by symmetric matrices is that they possess real eigenvalues whose eigenvectors form an orthogonal basis \cite{bhatia2013matrix}. Although L itself is not symmetric in general, it is similar to the symmetric matrix $O = B^{-1/2} SB^{-1/2}$ since
$$L=B^{-1}S=B^{-1/2}B^{-1/2}SB^{-1/2}B^{1/2}=B^{-1/2}OB^{1/2}$$
Thus, $L$ and $O$ have the same real eigenvalues \cite{zhang2010spectral}. Further, these eigenvalues can be compared for shape identification.

\begin{figure*}[t]
    \noindent
    \centering{
    \includegraphics[scale=0.33]{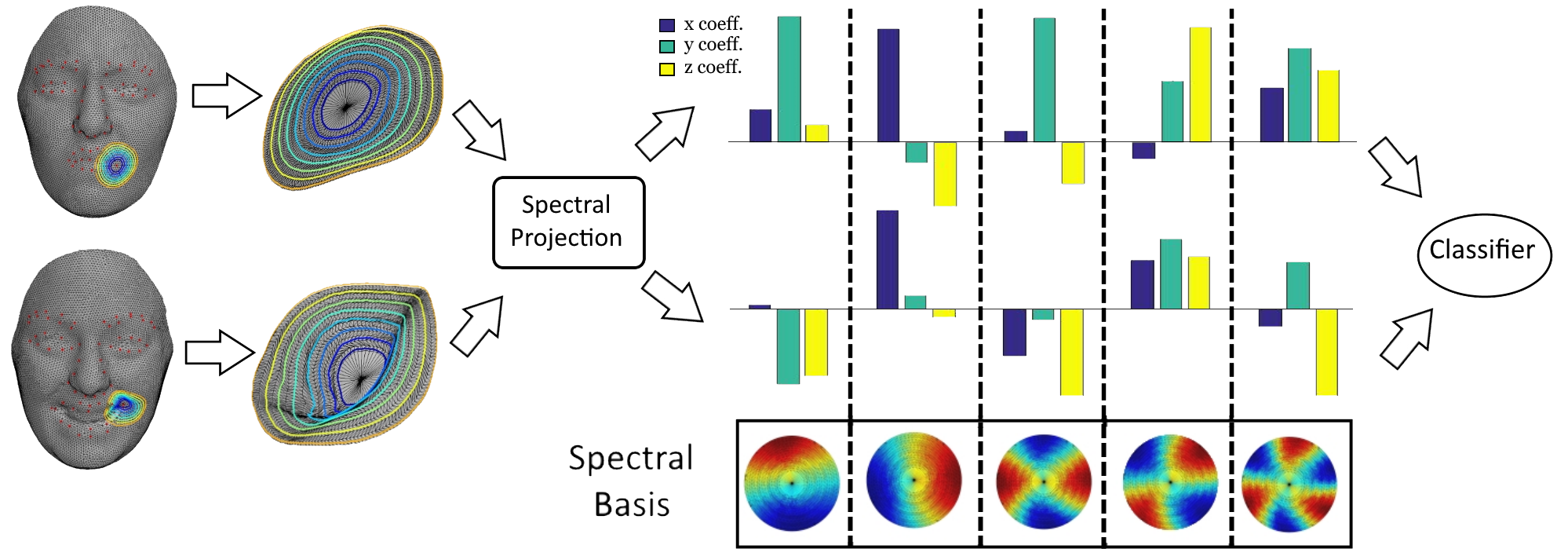} 
    }
    \caption{Schematic representation of the proposed approach. For each facial landmark, a surface patch is extracted to describe its local geometry. Each patch is projected into a common eigenspace to obtain a set of spectral coefficients that constitute our features. The eigenspace is computed off-line as the spectrum of the Graph Laplacian operator which depends exclusively on the connectivity of vertices and is therefore common for all patches. The spectral coefficients can be interpreted as loadings that weight the contribution of the spectral components. In the figure we display the coefficients of the first $5$ spectral components, as well as the spatial patterns produced by their corresponding eigenvectors.}
    \label{flowchart}
\end{figure*}

\section{SPECTRAL REPRESENTATION OF FACIAL PATCHES}

In order to explore the use of spectral methods as local shape descriptors, we represented a surface based on surface patches.
For this purpose, we choose to consider $N$ reference points (landmarks) $\lbrace r_{l}\rbrace_{1\le l \le N}$ (Fig. 1(a)) and, following \cite{maalej2011shape}, their associated sets of level curves $\lbrace c_{\lambda}^{l}\rbrace _{\lambda_{min} \le \lambda \le \lambda_{max} }$ (Fig. 1(b)). These curves were extracted over the surface $S$ centered at the considered landmark points, where $\lambda$ is the distance between the reference point $r_{l}$ and the point belonging to the curve $c_{\lambda}^{l}$, $\lambda_{min}$ and $\lambda_{max}$ stand for the minimum and maximum values taken by $\lambda$. The computation of the curves was performed using an Euclidean distance function:
$$c_{\lambda}^{l}=\lbrace p\in S | \| r_{l}-p  \| = \lambda \rbrace \subset S, \lambda \in [\lambda_{min}, \lambda_{max}] \eqno(3)$$
In that way, $c_{\lambda}^{l}$ is a curve, which consists of a collection of points $\lbrace p \rbrace $ located at an equal distance $\lambda$ from point $r_{l}$.
Accordingly, each facial surface is represented by patches that consist of sets of level curves around landmarks.

Once the patches are extracted, we aim to study their shape. Because we want to calculate the mesh spectra for the patches, we need to convert level curves to surface patches. Notice that, conceptually, we may directly extract the patches with no need to first extract the curves, but proceeding this way facilitates comparison to \cite{maalej2011shape} and, as we explain below, allows for using directly the graph Laplacian instead of the Laplace-Beltrami operator. To generate the mesh patches we re-sample the curves uniformly (as done in \cite{maalej2011shape}) and define a unique connectivity between them, which will be shared by all patches  (Fig. 1(d)).

After these pre-processing steps, we extract spectral features for facial expression analysis. We propose to do so using the Graph Laplacian, since this is the more theoretically sound approach under our settings. We also compare the results obtained by Shape-DNA, arguably the most widespread method to extract spectral features from 3D meshes. Specifically, spectral features are extracted as follows:
\begin{itemize}
    \item Graph Laplacian: Whereas graph Laplacian depends only on the connectivity between vertices, we calculated matrix $L$ using formula (1) only once. Eigenvalues and eigenvectors were obtained from this matrix. Because we generated all our mesh patches with the same order of connectivity, the set of eigenvectors constitutes a common basis to represent the spatial spectrum of all patches. Therefore, we used these eigenvectors to project mesh coordinates into the common eigenspace. These projections constitute our feature vectors, and are directly comparable between patches.
    \item Shape-DNA: The second type of features was obtained using the Laplace-Beltrami operator (2). This operator was calculated separately on each mesh-patch, because it depends not only on the connectivity but also on the location of the vertices. Thus, the eigen-decomposition of each patch produces a different eigenspace, which is tuned to the geometry of that specific patch. Projections into the eigenspace are therefore no longer comparable, but the eigenvalues resulting from each decomposition have been proven discriminative \cite{reuter2010hierarchical}, hence we use them as feature vectors.
\end{itemize}

To drive the classification experiments, we employed two different classifiers --- support vector machines (SVM) invoking the LIBSVM software \cite{chang2011libsvm} and Fisher's Linear Discriminant Analysis (FLDA) \cite{welling2005fisher}. A schematic diagram of the proposed framework is presented in Fig. \ref{flowchart}.

\section{EXPERIMENTS}

In the following section, we provide the details of our experiments on feature extraction with the proposed spectral analysis for facial expression recognition.

\subsection{Experimental setting}

In order to evaluate the proposed local shape spectrum analysis, we use the BU-3DFE database \cite{yin20063d}, which is one of the most widely used corpora for facial expression analysis in 3D. This database consist of 3D face scans of 100 subjects with different facial expressions. There are also variations in race, gender and age. Scans are annotated according  to the six prototypical facial expressions (anger, disgust, fear, happiness, sadness and surprise) at four different intensity levels. For our experiments, we have used the scans from all 100 subjects at the two highest intensity levels. Thus, our dataset consists of 1200 3D face scans, namely two intensity levels for each of the six facial expressions from 100 subjects.

Accompanying each facial scan there are 83 manually labeled landmarks. From these, 15 landmarks correspond to the silhouette contour and have arguably little validity in a 3D setting, hence we considered the subset of $N=68$ landmarks laying within the face area. All facial scans were represented by 68 patches $\lbrace c_{\lambda}^{l}\rbrace _{\lambda_{min} \le \lambda \le \lambda_{max} }$, Where, each patch consist of 15 level curves (Fig. 1(c)) ($\lambda_{min}=5, \lambda_{max}=20$) and each curve is a collection of points situated at an equal distance from the considered landmarks.


The dataset was arbitrarily divided into ten identity-disjoint sets; each of these (composed on 120 samples) was tested with models trained from the remaining nine sets (1080 samples). Thus, the recognition rates are obtained by averaging the results over the 10 sets (10-fold cross-validation).   
\begin{table}[t]
\caption{average accuracy of the three methods using two classifiers}
\label{t1}
\begin{center}
    \begin{tabular}{c||c c c }
    \hline
        & Curves & Graph Laplacian & Shape-DNA \\
    \hline
    \hline
    FLDA & 77.53\% & 81\% & 73.5\% \\
    SVM & 78.2\% & 81.5\% & 73.62\% \\
    \hline
    \end{tabular}
\end{center}
\end{table}

\begin{table}[t]
\caption{average confusion matrix of \textbf{Graph Laplacian} using 50 eigenvalues and an SVM classifier}
\label{confusionGL}
\begin{center}
    \begin{tabular}{c|| c c c c c c}
    \hline
     \% & AN & DI & FE & HA & SA & SU\\
    \hline
    \hline
     \rowcolor{gray!10}
    AN & \textbf{85.58} & 4.14 & 1.26 & 0.5 & 8.52 & 0\\ 
    DI & 7.5 & \textbf{75.31} & 8.76 & 3.01 & 0.9 & 4.52\\
     \rowcolor{gray!10}
    FE & 5.58 & 8.6 & \textbf{65.12} & 12.55 & 2.59 & 5.56 \\
    HA & 0 & 2.16 & 6.87 & \textbf{89.5} & 0 & 0.9\\
     \rowcolor{gray!10}
    SA & 14.5 & 0.76 & 7.46 & 0 & \textbf{77.2} & 0\\
    SU & 0 & 1.72 & 3.53 & 1.2 & 0 & \textbf{93.5}\\
    \hline
    \end{tabular}
\end{center}
\end{table}

\addtolength{\textheight}{+0cm}   
\subsection{Results on Expression Recognition}

\begin{table}[t]
\caption{average confusion matrix of \textbf{Shape-DNA} using 50 eigenvalues and an SVM classifier}
\label{confusionShapeDNA}
\begin{center}
    \begin{tabular}{c||c c c c c c }
    \hline
     \% & AN & DI & FE & HA & SA & SU\\
    \hline
     \hline
      \rowcolor{gray!10}
    AN & \textbf{77.21} & 5.87 & 2.71 & 1.21 & 12.98 & 0\\
    DI & 7.45 & \textbf{75.53} & 7.98 & 3.87 & 1.52 & 3.61\\
     \rowcolor{gray!10}
    FE & 7.23 & 9.82 & \textbf{52.53} & 15.46 & 9.56 & 5.37 \\
    HA & 2.1 & 3.45 & 12.5 & \textbf{80.49} & 0 & 1.46\\
     \rowcolor{gray!10}
    SA & 19.76 & 3.51 & 6.32 & 0.31 & \textbf{69.51} & 0.57\\
    SU & 0.49 & 1.52 & 10.23 & 1.54 & 0.49 & \textbf{85.74}\\
    \hline
    \end{tabular}
\end{center}
\end{table}

\begin{table}[t]
\caption{average confusion matrix of approach based on \textbf{"Distances between curves "} using an SVM classifier}
\label{confusionCurves}
\begin{center}
    \begin{tabular}{c||cccccc}
    \hline
     \% & AN & DI & FE & HA & SA & SU\\
    \hline
    \hline
     \rowcolor{gray!10}
    AN & \textbf{78.96} & 6.08 & 3.83 & 0.55 & 10.56 & 0\\
    DI & 5.49 & \textbf{76.14} & 5.09 & 4.42 & 2.87 & 5.97\\
     \rowcolor{gray!10}
    FE & 3.92 & 7.08 & \textbf{63.45} & 13.24 & 6.12 & 6.17 \\
    HA & 1.11 & 2.78 & 9.45 & \textbf{86.65} & 0 & 0\\
     \rowcolor{gray!10}
    SA & 12.26 & 0.58 & 8.28 & 0 & \textbf{78.86} & 0\\
    SU & 0 & 2.78 & 8.86 & 1.08 & 0.55 & \textbf{86.71}\\
    \hline
    \end{tabular}
\end{center}
\end{table}

Our first experiment consists on a direct comparison of the proposed spectral features (based on GLF) with respect to the curves-framework and with respect to Shape-DNA, which constitute the straight-forward spectral alternative. This was done in the context of expression recognitions targeting the six basic emotions. (anger (AN), disgust (DI), fear (FE), happiness (HA), sadness (SA), surprise (SU)). Table \ref{t1} 
summarizes the average accuracy obtained by each approach. It can be seen that the spectral features based on the Graph Laplacian outperform the curves-based approach, which suggest that they can capture a more complete information about the facial patches. It is also interesting to see that Shape-DNA features obtain the lowest accuracy among the three methods. This confirms the theoretical limitations already highlighted with respect to the direct application of Shape-DNA to surface patches: given two shapes to compare under a spectral representation, small differences between them can modify the eigen-decomposition to the extent that the eigenvalues change their relative order producing a swapping of the extracted basis \cite{jain2007non}. Such swaps make the direct comparison of eigenvalues used in Shape-DNA conceptually incorrect. Fixing this would require matching algorithms to appropriately re-order the resulting eigenvalues. Our GLF do not suffer from this issue as they result from a projection into a common basis, which only depends on the connectivity of the patches.


\begin{table*}[t!]
\caption{average accuracy of facial expression recognition under different classifiers for different numbers of eigenvalues}
\label{table1}
\centering
    \begin{tabular}{c||ccccc||ccccc}
    \hline
    Features & \multicolumn{5}{c||}{Graph Laplacian} & \multicolumn{5}{c}{Shape-DNA}\\
    \hline
  Eigenvalues & 200 & 100 & 50 & 30 & 10 & 200 & 100 & 50 & 30 & 10 \\
    \hline
    FLDA & 80.25\% & 79.92\% & \textbf{81}\% & 80.25\% & 79.42\% & 71.17\% & 71.25\% & \textbf{73.5}\% & 72.83\% & 71.08\% \\
  SVM & 80.3\% & 80\% & \textbf{81.5}\% & 79.5\% & 80.83\% & 71.2\% & 71.33\% & \textbf{73.62}\% & 72.9\% & 71\% \\
    \hline
    \end{tabular}
\end{table*}


To put our results in a wider context, we can also compare them to other methods reporting expression recognition rates on the BU3DFE database. As detailed in \cite{zeng2013automatic}, only methods whose experimental settings consider the whole set of 100 subjects are fairly comparable. Among these, expression recognition rates vary between $68.2\%$ \cite{zeng2013automatic} and $82.7\%$ \cite{yang2015automatic}, while our average recognition rates reach $81.5\%$. Notice that in our case we use a single type of feature (GLF), while most other works achieving high recognition rates use combinations of multiple features.

To provide a more extensive review of our results, Tables \ref{confusionGL}, \ref{confusionShapeDNA} and \ref{confusionCurves} show the average confusion matrices for each of the approaches using SVM classifier. It can be seen that among the six basic expressions surprise, happiness and anger were recognized the best. In contrast, fear and disgust were the most difficult expressions to predict. We also observe that GLF consistently outperform both the curve-based and Shape-DNA approaches for most expressions, with the only major exception being Disgust, where it performs similarly but slightly worse than the competing alternatives. 

An important factor when using spectral decomposition methods is the number of considered components. All results reported above correspond to the first 50 components (eigenvalues in the case of Shape-DNA, projections in the eigen-space in the case of GLF). We also repeated the expression recognition experiments for different numbers of components and found the performance of both GLF and Shape-DNA to be relatively as long as at least 10 components were used (see Table \ref{table1}). Tests extended only up to 200 components, since increasing the components implies also more computational load while not bringing improvements in accuracy.

\subsection{Action Unit Estimation}

Since our approach is based on the aggregation of localized descriptors of the facial surface, it would make sense that it can also be applied to the estimation of Action Units (AU). AUs are designed to capture any anatomically feasible facial deformation \cite{ekman1994strong}, thereby combinations of AUs can be used to describe any of the six basic expressions \cite{ruiz2015emotions}, as well as any other anatomically feasible facial expression. Each of the expressions in the BU-3DFE database was manually annotated with corresponding sets of AUs by two coders\footnote{Available at http://fsukno.atspace.eu/Research.htm\#FG2017a}. The resulting annotations were checked for consistency of the obtained AU frequencies per expression and co-occurrences of AUs with \cite{du2014compound, wang2014capture, zhao2015joint}. Then, experiments on AUs recognition were performed under the same conditions as the expressions recognition tests. 

Table \ref{F1scoreAU} shows the weighted average F1-score for each AU (weighted proportionally to the number of samples per AU). One common characteristic of all the approaches is that they all recognized AU25, AU26 better that any other. Also, analysing the table, we can see that detection of AU1, AU2, AU4, AU5 and AU12 can be said reliable. The worst detected AU was AU15. 

When comparing among features, our results show the same tendency observed in the expression recognition experiments. The best performance was obtained by GLF, which clearly outperformed Shape-DNA and was also slightly better than the approach based on geodesic distance between curves. Regarding the latter, while the average recognition accuracy of GLF and curves were rather similar, it should be noted that GLF consistently outperformed curves in 15 out of the 17 tested AUs. 


\section{CONCLUSIONS}

In this work, we extend the analysis of 3D geometry from a curve-based representation into a  spectral  representation. This representation allows to build a complete description of the underlying surface that can be further tuned to the desired level of detail. We propose the use of Graph Laplacian Features (GLF), which result from the projection of local surface patches into a common basis obtained from the Graph Laplacian eigenspace, much like a Fourier transform into the spatial frequency bassis of the surface patches. Further, we compare our approach with two others approaches. The first one is the curves-based framework and the second one is the straight-forward alternative for spectral representation, Shape-DNA, which is based on the Laplace Beltrami Operator. We show that the straight-forward application of Shape-DNA is not the best way to deal with local face patches, since it cannot provide a stable basis to guarantee that the extracted signatures for the different patches are directly comparable.

We tested the proposed approach in the BU-3DFE database in terms of expressions and Action Units recognition. Our results show that the proposed GLF consistently outperform the curves-based and Shape-DNA alternatives, both in terms of expression recognition and Action Unit recognition. Moreover, the recognition rates of shape-DNA are even lower than the curves-based framework, as predicted by the theory: in spite of upgrading the curves-based representation to a full-surface description, similarly to GLF, the instabilities of the bases extracted by Shape-DNA result in a decreased performance.

Interestingly, the accuracy improvement brought by GLF is obtained also at a lower computational cost. Considering the extraction of patches as a common step between the three compared approaches, the curve-based framework requires a costly elastic deformation between corresponding curves (e.g. based on splines) and Shape-DNA requires computing the eigen-decomposition of each new patch to be analyzed. In contrast, GLF only require the projection of the patch geometry into the Graph Laplacian eigenspace, which is common to all patches and can thus be pre-computed off-line. 

\begin{table}[t]
\caption{average F1-score recognition results of AUs using 50 eigenvalues}
\label{F1scoreAU}
\begin{center}
    \begin{tabular}{c||c||ccc}
     
    \hline
     AU & \# Samples & Curves & Graph Laplacian & Shape-DNA \\
    \hline
    \hline
    \rowcolor{gray!10}
    1 & 333 & 0.74 & 0.75 & 0.73 \\

    2 & 302 & 0.77 & 0.78 & 0.73 \\
    \rowcolor{gray!10}
    4 & 423 & 0.77 & 0.79 & 0.74 \\
    5 & 304 & 0.76 & 0.80 & 0.71 \\
    \rowcolor{gray!10}
    6 & 68 & 0.42 & 0.46 & 0.45 \\
    7 & 370 & 0.69 & 0.73 & 0.63 \\
    \rowcolor{gray!10}
    9 & 99 & 0.55 & 0.56 & 0.47\\
    10 & 136 & 0.64 & 0.67 & 0.57 \\
   \rowcolor{gray!10}
    12 & 177 & 0.74 & 0.76 & 0.70 \\
    15 & 69 & 0.37 & 0.34 & 0.30 \\
    \rowcolor{gray!10}
    16 & 122 & 0.50 & 0.52 & 0.39 \\
    17 & 130 & 0.48 & 0.50 & 0.42 \\
    \rowcolor{gray!10}
    20 & 84 & 0.28 & 0.3 &  0.25\\
    23 & 134 & 0.42 & 0.50 & 0.38 \\
    \rowcolor{gray!10}
    24 & 125 & 0.57 & 0.61 & 0.62 \\
    25 & 709 & 0.94 & 0.94 & 0.92 \\
    \rowcolor{gray!10}
    26 & 230 & 0.85 & 0.88 & 0.86 \\
    \hline
    \textbf{Avrg} & Total: 3815 & \textbf{0.72} & \textbf{0.74} & \textbf{0.69}\\
     \hline
    \end{tabular}
\end{center}
\end{table}

\addtolength{\textheight}{+0.2cm}
\section*{ACKNOWLEDGMENTS}

This work is partly supported by the Spanish Ministry of Economy and Competitiveness under the Ramon y Cajal fellowships and the Maria de Maeztu Units of Excellence Programme (MDM-2015-0502).



\bibliographystyle{IEEEtran}
\bibliography{egbib}



\end{document}